\documentclass[nohyperref]{article}

\usepackage{microtype}
\usepackage{graphicx}
\usepackage{booktabs} % for professional tables
\usepackage{caption}
\usepackage{subcaption}
\usepackage{makecell}

\usepackage{hyperref}
\usepackage{xcolor}
\usepackage{amsmath,bm,multicol,multirow}
\usepackage{pgfplotstable}
\usepackage[scientific-notation=true]{siunitx}

\usepgfplotslibrary{fillbetween}
\usetikzlibrary[patterns]
\usetikzlibrary{shapes.geometric}
\usepackage[accepted]{icml2023}

\usepackage{amssymb}
\usepackage{mathtools}
\usepackage{amsthm}

\usepackage[capitalize,noabbrev]{cleveref}

\theoremstyle{plain}

\theoremstyle{definition}

\theoremstyle{remark}

\usepackage[textsize=tiny]{todonotes}

\newcommand{\ma}[1]{{\color{orange}#1}}

\renewcommand{\ma}[1]{}  %uncomment to hide marks
\newcommand{\name}{data2vec 2.0}

\newcommand{\wvpp}{wav2vec 2.0}

\newcommand{\libri}{Librispeech}
\newcommand{\libril}{Libri-light}
\newcommand{\voxsz}{LL-60K}
\newcommand{\librisz}{LS-960}

\newcommand{\Enot}[1]{\num[exponent-product = \times]{#1}}
\newcommand{\insertEfficiencyPlot}{

\pgfplotstableread[row sep=\\,col sep=&]{
epochs & updates & accuracy & hours\\
200 & 500000 & 84.5 & 32.0 \\
150 & 375300 & 84.42 & 23.97\\
40 & 100000 & 83.87 & 4.65\\
20 & 50000 & 83.7 & 3.09\\
10 & 25000 & 81.89 & 1.3\\
}\visionTrainingTimeDataToVecTwo

\pgfplotstableread[row sep=\\,col sep=&]{
epochs & updates & accuracy & hours\\
300	& 187500 & 84.0 & 17.81\\
800	& 500000 & 84.2 & 47.5\\
}\visionTrainingTimeDataToVec

\pgfplotstableread[row sep=\\,col sep=&]{
epochs & updates & accuracy & hours\\
1600 & 500000 & 83.6 & 50.7\\
}\visionTrainingTimeMAE

\pgfplotstableread[row sep=\\,col sep=&]{
upd & wer & trhours \\
400000 & 10.9 & 43.3 \\ %final based on 50k updates
300000 & 11.5 & 32.47 \\
200000 & 11.9 & 21.65 \\
100000 & 12.8 & 10.82 \\
50000  & 16.0 & 5.41 \\
}\speechTrainingTimeDataToVecTwo

\pgfplotstableread[row sep=\\,col sep=&]{
upd & wer & trhours \\
400000 & 12.0 & 63.33 \\ %final based on 10k updates
}\speechTrainingTimeDataToVec

\pgfplotstableread[row sep=\\,col sep=&]{
upd & wer & trhours \\
400000 & 16.1 & 57.27 \\ %final based on 50k updates
}\speechTrainingTimeWavToVecTwo

\pgfplotstableread[row sep=\\,col sep=&]{
upd & glue & trhours & ep\\
1000000 & 82.6 & 28.2 & 4.1\\
900000 & 82.5 & 25.4 & 3.7 \\ 
800000 & 82.3 & 23.8 & 3.3 \\
600000 & 81.6 & 17.0 & 2.4 \\
400000 & 80.8 & 11.4 & 1.6 \\
}\nlpTrainingTimeDataToVecTwo

\pgfplotstableread[row sep=\\,col sep=&]{
upd & glue & trhours & ep \\
1000000 & 82.7 & 69.4 & 31.8 \\ %final based on 10k updates
}\nlpTrainingTimeDataToVec

\pgfplotstableread[row sep=\\,col sep=&]{
upd & glue & trhours & ep\\
1000000 & 82.5 & 50.5 & 31.8 \\ %final based on 50k updates
}\nlpTrainingTimeRoBERTa

\begin{figure*}[h!]
\centering
\begin{subfigure}[b]{.37\textwidth}
\begin{tikzpicture}
\begin{axis}[
width=0.9\columnwidth,
height=.7\columnwidth,
legend style={font=\small,%fill=none,
at={(0.98,0.27)},
anchor=east,legend columns=1,draw=none},
legend cell align={left},
xticklabel style={font=\small},
yticklabel style={font=\small},
ymin=81,ymax=85,
ylabel={Top-1 Accuracy},
ylabel near ticks, %ylabel shift={-10pt}
ylabel style={font=\small},
xlabel={Training Time (hours)},
xlabel near ticks,
xlabel style={font=\small},
grid=both,
]
\addplot[<->,dashed,black,forget plot] coordinates {(5,83.6)(48.5,83.6)};
\node[font=\scriptsize] at (axis cs: 27,83.4) {16.4x};
\addplot[blue,mark=square*,mark options={solid,scale=1,fill=blue}] table[x=hours,y=accuracy]{\visionTrainingTimeDataToVecTwo};
\addplot[red,mark=*,mark options={solid,scale=1.25,fill=red}] table[x=hours,y=accuracy]{\visionTrainingTimeDataToVec};
\addplot[orange,mark=triangle*,mark options={solid,scale=1.5,fill=orange}] table[x=hours,y=accuracy]{\visionTrainingTimeMAE};
\legend{data2vec 2.0,data2vec,MAE}
\end{axis}
\end{tikzpicture}
\caption{Computer Vision.}
\label{fig:efficiency_vision}
\end{subfigure}
\hspace{-12mm}
\begin{subfigure}[b]{.37\textwidth}
\begin{tikzpicture}
\begin{axis}[
width=0.9\columnwidth,
height=.7\columnwidth,
legend style={font=\small,%fill=none,
at={(0.88,0.57)},
anchor=east,legend columns=1,draw=none},
legend cell align={left},
xticklabel style={font=\small},
yticklabel style={font=\small},
ymin=10,ymax=17,
ylabel={Word Error Rate},
ylabel near ticks, %ylabel shift={-10pt}
ylabel style={font=\small},
xlabel={Training Time (hours)},
xlabel near ticks,
xlabel style={font=\small},
grid=both,
]
\addplot[<->,dashed,black,forget plot] coordinates {(7.3,16.0)(54.7,16.0)};
\node[font=\scriptsize] at (axis cs: 31,16.4) {10.6x};
\addplot[blue,mark=square*,mark options={solid,scale=1,fill=blue}] table[x=trhours,y=wer]{\speechTrainingTimeDataToVecTwo};
\addplot[red,mark=*,mark options={solid,scale=1.25,fill=red}] table[x=trhours,y=wer]{\speechTrainingTimeDataToVec};
\addplot[teal,mark=square*,mark options={solid,scale=1.25,fill=teal}] table[x=trhours,y=wer]{\speechTrainingTimeWavToVecTwo};
\legend{data2vec 2.0, data2vec, wav2vec 2.0}
\end{axis}
\end{tikzpicture}
\caption{Speech Processing.
}
\label{fig:efficiency_speech}
\end{subfigure}
\hspace{-12mm}
\begin{subfigure}[b]{.37\textwidth}
\begin{tikzpicture}
\begin{axis}[
width=0.9\columnwidth,
height=.7\columnwidth,
legend style={font=\small,%fill=none,
at={(0.98,0.27)},
anchor=east,legend columns=1,draw=none},
legend cell align={left},
xticklabel style={font=\small},
yticklabel style={font=\small},
ymin=80.0,ymax=83.5,
ylabel={GLUE score},
ylabel near ticks, %ylabel shift={-10pt}
ylabel style={font=\small},
xlabel={Training Time (hours)},
xlabel near ticks,
xlabel style={font=\small},
grid=both,
]
\addplot[<->,dashed,black,forget plot] coordinates {(27.3,82.5)(48.7,82.5)};
\node[font=\scriptsize] at (axis cs: 40.0,82.7) {2x};
\addplot[blue,mark=square*,mark options={solid,scale=1,fill=blue}] table[x=trhours,y=glue]{\nlpTrainingTimeDataToVecTwo};
\addplot[red,mark=*,mark options={solid,scale=1.25,fill=red}] table[x=trhours,y=glue]{\nlpTrainingTimeDataToVec};
\addplot[olive,mark=diamond*,mark options={solid,scale=1.5,fill=olive}] table[x=trhours,y=glue]{\nlpTrainingTimeRoBERTa};
\legend{data2vec 2.0, data2vec, RoBERTa}
\end{axis}
\end{tikzpicture}
\caption{Natural Language Processing.
}
\label{fig:efficiency_nlp}
\end{subfigure}
\caption{Efficiency of \name{} for computer vision and speech processing in terms of wall clock time for pre-training Base models.
Vision models are pre-trained on ImageNet-1K using 32 A100 40GB GPUs, then fine-tuned to perform image recognition and we report top-1 dev accuracy.
Pre-training of speech models uses Librispeech and 16 A100 40GB GPUs, models are fine-tuned for speech recognition on the 10 hour labeled data split of Libri-light and we report word error rate on dev-other without a language model.
}
\label{fig:efficiency}
\end{figure*}

}

\newcommand{\insertINtable}{
\begin{table}[t]
\centering
\caption{Computer vision: top-1 validation accuracy on ImageNet-1K for ViT-B and ViT-L. 
\label{tab:in}}
\begin{tabular}[t]{p{4cm}rrr}
\toprule
& epochs & ViT-B & ViT-L \\
\midrule
\multicolumn{4}{l}{\textit{Multiple models/external data}} \\
BEiT~\citep{bao2021beit} & 800 & 83.2 & 85.2 \\
PeCo~\citep{dong2022peco} & 800 & 84.5 & 86.5 \\
BEiT-2~\citep{peng2022beit2} & 1600 & 85.5 & 87.3 \\
TEC~\citep{gao2022sustainssl} &  \thead[r]{2400/\\1900} & 85.1 & 86.5 \\
\midrule
\textit{Single models} \\
MoCo-3~\citep{chen2021mocov3} & 300 & 83.2 & 84.1 \\
DINO~\citep{caron2021dino} & 1600 & 82.8 & - \\
MAE~\citep{he2021mae} & 1600 & 83.6 & 85.9 \\
SimMIM~\citep{xie2021simmim} & 800 & 83.8 & - \\
iBOT~\citep{zhou2021ibot} & 1600 & 83.8 & - \\
MaskFeat~\citep{wei2021masked} & 1600 & 84.0 & 85.7 \\
data2vec~\citep{baevski2022d2v} & \thead[r]{800/\\1600} & 84.2 & 86.6 \\
\name{} & \thead[r]{200/\\150} & 84.5 & 86.8 \\
\bottomrule
\end{tabular}
\end{table}

\begin{table}[t]
\centering
\caption{Computer vision: top-1 validation accuracy on ImageNet-1K for ViT-H/14. Pre-training time measurd on 64 A100 GPUs.
\label{tab:inhuge}}
\begin{tabular}[t]{lrrr}
\toprule
& epochs & ViT-H & Pre-train \\
& & & time (h) \\
\midrule
MAE~\citep{he2021mae} & 1600 & 86.9 & 113.6 \\
\name{} & 100 & 87.4 & 66.1 \\
\bottomrule
\end{tabular}
\end{table}
}

\newcommand{\insertLLtable}{
\begin{table*}[h!]
\caption{Speech processing: word error rate on the \libri{} test-other when fine-tuning pre-trained models on the \libril{} low-resource labeled data setups~\citep{kahn2020librilight} of 10 min, 1 hour, 10 hours, the clean 100h subset of \libri{} and the full 960h of \libri{}. 
For pretraining, models use 960 hours of unlabeled audio from \libri{} (\librisz{}), or the 60K hours from \libril{} (\voxsz{}); WavLM Large uses 94K hours (MIX-94K) which includes \voxsz{} as well as other datasets.
All results are based on 4-gram language models.
We report wall clock pre-training time for Base models on 16 A100 GPUs and Large models on 64 A100 GPUs.
\label{tab:librilight}
}
\vspace{0.075in}
\centering 
\begin{tabular}{lccrrrrrr}
\toprule
& Unlabeled & \multirow{2}{*}{LM} & \multicolumn{5}{c}{Amount of labeled data} & Pre-train \\
& data & {} & 10m & 1h & 10h & 100h & 960h & time (h)\\
\midrule
\textit{Base models} \\
\wvpp{}~\citep{baevski2020wav} & \librisz{} & 4-gram & 15.6 & 11.3 & 9.5 & 8.0 & 6.1 & 57.3 \\
HuBERT~\citep{hsu2020hubert} & \librisz{} & 4-gram & 15.3 & 11.3 & 9.4 & 8.1 & - & - \\
WavLM~\citep{chen2021wavlm} & \librisz{} & 4-gram & - & 10.8 & 9.2 & 7.7 & - & - \\
data2vec~\citep{baevski2022d2v} & \librisz{} & 4-gram & 12.3 & 9.1 & 8.1 & 6.8 & 5.5 & 63.3 \\
\name{}  & \librisz{} & 4-gram & 11.5 & 8.7 & 7.6 & 6.4 & 5.2 & 43.3 \\
\midrule
\textit{Large models} \\
\wvpp{}~\citep{baevski2020wav} & \voxsz{} & 4-gram & 10.3 & 7.1 & 5.8 & 4.6 & 3.6 & 150.0 \\ %10,240\\
HuBERT~\citep{hsu2020hubert} & \voxsz{} & 4-gram & 10.1 & 6.8 & 5.5 & 4.5 & 3.7 & - \\
WavLM~\citep{chen2021wavlm} & MIX-94K & 4-gram & - & 6.6 & 5.5 & 4.6 & - & - \\
data2vec~\citep{baevski2022d2v} & \voxsz{} & 4-gram & 8.4 & 6.3 & 5.3 & 4.6 & 3.7 & 108.0 \\ % % 6,912\\
\name{} & \voxsz{} & 4-gram & 8.4 & 6.3 & 5.1 & 4.3 & 3.5 & 76.7 \\ % 4,907\\
\bottomrule
\end{tabular}
\end{table*}
}

\newcommand{\insertGLUEtable}{
\begin{table*}[h!]
\centering
\caption{Natural language processing: GLUE results on the dev set for single-task fine-tuning with Base models.
For MNLI we report accuracy on the matched/unmatched dev sets, for MRPC and QQP, we report the unweighted average of accuracy and F1, for STS-B the unweighted average of Pearson and Spearman correlation, for CoLA Matthews correlation and accuracy for all other tasks.
BERT Base results are from~\citet{wu2020clear}, the baseline is a reproduction of BERT, pre-training time (PT) is measured on 16 A100 GPUs.
\label{tab:glue}
} 
\vspace{0.075in}
\begin{tabular}{lrrrrrrrrrrr}
\toprule
& epochs & MNLI & QNLI & RTE & MRPC & QQP & STS-B & CoLA & SST & Avg. & Pre-train \\
& & & & & & & & & & & time (h) \\
\midrule
BERT & - & 84.0/84.4 & 89.0 & 61.0 & 86.3 & 89.1 & 89.5 & 57.3 & 93.0 & 81.2 & - \\
Baseline & 31.8 & 84.1/83.9 & 90.4 & 69.3 & 89.0 & 89.3 & 88.9 & 56.8 & 92.3 & 82.5 & 50.5 \\ 
data2vec & 31.8 & 83.2/83.0 & 90.9 & 67.0 & 90.2 & 89.1 & 87.2 & 62.2 & 91.8 & 82.7 & 69.4 \\
\name{} & 4.1 & 83.7/83.7 & 90.7 & 68.6 & 88.8 & 89.3 & 87.3 & 59.1 & 92.9 & 82.6 & 28.2 \\
\bottomrule
\end{tabular}
\end{table*}
}

\newcommand{\insertBszCloneAblationPlot}{

\pgfplotstableread[row sep=\\,col sep=&]{
bsz & clone & updates & epochs &   accuracy & hours\\
2 & 2 & 100k & 6 & 77.8 & 3.15\\
2 & 4 & 100k & 6 & 79.85 & 3.25\\
2 & 8 & 100k & 6 & 81.17 & 3.12\\
2 & 16 & 100k & 6 & 82.36 & 3.2\\
}\bszCloneAblationbTwo

\pgfplotstableread[row sep=\\,col sep=&]{
bsz & clone & updates & epochs &  accuracy & hours\\
4 & 1 & 100k & xx & 76.82 & xx\\
4 & 2 & 100k & 11 & 79.98 & 3.39\\
4 & 4 & 100k & 11 & 81.3 & 3.39\\
4 & 8 & 100k & 11 & 82.5 & 3.45\\
4 & 16 & 100k & 11 & 83.18 & 3.84\\
}\bszCloneAblationbFour

\pgfplotstableread[row sep=\\,col sep=&]{
bsz & clone & updates & epochs &  accuracy & hours\\
8 & 1 & 100k & xx & 79.09 & xx\\
8 & 2 & 100k & 21 & 81.62 & 3.87\\
8 & 4 & 100k & 21 & 82.72 & 4.01\\
8 & 8 & 100k & 21 & 83.58 & 4.43\\
8 & 16 & 100k & 21 & 83.85 & 4.72\\
}\bszCloneAblationbEight

\pgfplotstableread[row sep=\\,col sep=&]{
bsz & clone & updates & epochs &  accuracy &hours\\
16 & 1 & 100k & 41 & 80.99 & 4.92\\
16 & 2 & 100k & 41 & 83.07 & 4.92\\
16 & 4 & 100k & 41 & 83.64 & 5.16\\
16 & 8 & 100k & 41 & 83.86 & 5.25\\
16 & 16 & 100k & 41 & 84.23 & 7.12\\
}\bszCloneAblationbSixteen

\pgfplotstableread[row sep=\\,col sep=&]{
bsz & clone & updates & epochs &  accuracy &hours\\
32 & 1 & 100k & 81 & 82.31 & x\\
32 & 4 & 100k & 81 & 83.89 & 6.98\\
32 & 8 & 100k & 81 & 84.16 & 6.96\\
32 & 16 & 100k & 81 & 84.18 & 11.68\\
}\bszCloneAblationbThirtyTwo

\begin{figure}
\centering
\begin{tikzpicture}
\begin{axis}[
width=.9\columnwidth,
height=.7\columnwidth,
xmode=log,
log basis x={2},
log ticks with fixed point,
legend style={font=\small,
at={(1.0,0.27)},
anchor=east,legend columns=1,draw=none},
legend cell align={left},
xtick={1,2,4,8,16,32},
ymin=74.5,
xmax=16,
ylabel={Top-1 Accuracy},
ylabel near ticks, %ylabel shift={-10pt}
xlabel={$M$},
grid=both,
]
\addplot[solid,blue,mark=*,mark options={solid,scale=1,fill=blue}] table[x=clone,y=accuracy]{\bszCloneAblationbTwo};
\addplot[dashed,green,mark=triangle*,mark options={solid,scale=1,fill=green}] table[x=clone,y=accuracy]{\bszCloneAblationbFour};
\addplot[dashdotdotted,black,mark=square*,mark options={solid,scale=1,fill=black}] table[x=clone,y=accuracy]{\bszCloneAblationbEight};
\addplot[densely dashdotted,orange,mark=diamond*,mark options={solid,scale=1,fill=orange}] table[x=clone,y=accuracy]{\bszCloneAblationbSixteen};
\addplot[solid,red,mark=otimes,mark options={solid,scale=1,fill=red}] table[x=clone,y=accuracy]{\bszCloneAblationbThirtyTwo};
\legend{bsz=64,bsz=128,bsz=256,bsz=512,bsz=1024}
\end{axis}
\end{tikzpicture}
\caption{Multi-mask training (\textsection\ref{sec:multimask}) enables pre-training with smaller batch sizes than usual. 
We show top-1 dev accuracy on ImageNet-1K for models pretrained with different batch sizes (bsz) and different number of masks per sample ($M$; \textsection\ref{sec:multimask}); bsz=64 and $M=1$ diverged due to too small overall batch size.
}
\label{fig:multimask}
\end{figure}
}

\newcommand{\insertDatasetSzPlot}{
\pgfplotstableread[row sep=\\,col sep=&]{
sz & vitb & vitl \\
100 & 84.43 & 86.5 \\
50 & 84.38 & 86.1 \\
25 & 84.24 & 85.85 \\
10 & 83.71 & 84.7 \\
5 & 82.94 & 82.6 \\
2.5 & 81.66 & \\
1 & 79.42 & 79.0 \\
}\datasetSzPlot

\begin{figure}[h!]
\centering
\begin{tikzpicture}
\begin{axis}[
width=.5\columnwidth,
height=.4\columnwidth,
xmode=log,
log basis x={10},
log ticks with fixed point,
legend style={%font=\small,
at={(0.98,0.17)},
anchor=east,legend columns=1,draw=none},
xtick={1,2.5,5,10,25,50,100},
ylabel={Top-1 Accuracy},
ylabel near ticks, %ylabel shift={-10pt}
xlabel={Amount of ImageNet-1K pre-training data (\%)},
grid=both,
]
\addplot[solid,black,mark=*,mark options={solid,scale=1,fill=blue}] table[x=sz,y=vitb]{\datasetSzPlot};
\addplot[solid,black,mark=*,mark options={solid,scale=1,fill=orange}] table[x=sz,y=vitl]{\datasetSzPlot};
\legend{\name{} ViT-B, \name{} ViT-L}
\end{axis}
\end{tikzpicture}
\caption{Top-1 accuracy when finetuning on the entire ImageNet-1K after pre-training on a subset of ImageNet-1K data.
}
\label{fig:datasetsize}
\end{figure}
}

\newcommand{\insertAblationTable}{

\begin{table*}[t]
\begin{minipage}{0.29\linewidth}
\begin{center}
\caption{Training losses. 
ImageNet accuracy for removing the CLS loss, adding pixel regression (pixel regr), and only pixel regression. 
Results use a reduced setup (100 epochs).
\label{tab:losses}}
\begin{tabular}[t]{lr}
\toprule
& top-1 (\%) \\
\midrule
baseline & 84.4 \\
- cls loss & 84.2 \\
+ pixel regr & 84.3 \\
pixel regr only & 83.5 \\
\multicolumn{2}{c}{~}\\
\bottomrule
\end{tabular}
\end{center}
\end{minipage}
\hspace{2em}
\begin{minipage}{0.29\linewidth}
\begin{center}
\caption{Masking strategy. Effect of block masking as well as different block sizes ($B$) for inverse block masking. $B=1$ is equivalent to random masking and inv. block $B=3$ is the default. %Setup follows~\autoref{tab:losses}.
\label{tab:masking}}
\begin{tabular}[t]{lr}
\toprule
& top-1 (\%) \\
\midrule
block $B=3$ & 84.1 \\
inv. block $B=1$ & 83.7 \\
inv. block $B=2$ & 84.4 \\
inv. block $B=3$ & 84.4 \\
inv. block $B=4$ & 84.2 \\
\bottomrule
\end{tabular}
\end{center}
\end{minipage}
\hspace{2em}
\begin{minipage}{0.29\linewidth}
\begin{center}
\caption{Alibi embeddings. WER on dev-other when removing Alibi, using only a single scalar for all heads and not learning scalars at all; pre-training is on LS-960 and fine-tuning with LL-10h.
\label{tab:alibi}}
\begin{tabular}[t]{lr}
\toprule
& WER \\
\midrule
baseline & 10.9 \\
- alibi & 11.3 \\
- learn scale/head & 11.0 \\
- learn scale & 11.7 \\
\multicolumn{2}{c}{~}\\
\bottomrule
\end{tabular}
\end{center}
\end{minipage}
\end{table*}
}

\newcommand{\insertHyperparameterTableVision}{
\begin{table}[h!]
\centering
\caption{Vision pre-training hyper-parameters. IN is instance normalization; AVG is mean pooling; LN is layer normalization.
\label{tab:imagenet_hyperparams}}
\begin{tabular}[t]{lrrr}
\toprule
& ViT-B & ViT-L & ViT-H/14 \\
\midrule
GPUs & 32 & 32 & 32 \\
Learning rate & $\Enot{1e-3}$ & $\Enot{4e-4}$& $\Enot{4e-4}$ \\
Adam $\beta_1$ / $\beta_2$ & 0.9 / 0.95 & 0.9 / 0.95& 0.9 / 0.95 \\
Weight decay & 0.05 & 0.05& 0.05 \\
Clip norm & 4.0 & 4.0& 4.0 \\
Learning rate schedule & cosine & cosine& cosine \\
Warmup updates & 50,040 & 50,040& 50,040 \\
Batch size (per GPU / total) & 16 / 512 & 8 / 256& 8 / 256 \\
Multi-masks ($M$) & 16 & 16& 16 \\
CLS loss coefficient & 0.01 & 0.01& 0.01 \\
$\tau_0$ (EMA start) & 0.9998 & 0.9998& 0.9998 \\ 
$\tau_e$ (EMA end) & 0.99999 & 1.0& 1.0 \\ 
$\tau_n$ (EMA anneal steps) & 100,000 & 500,000& 300,000 \\ 
$B$ (block width) & 3 & 3& 3 \\
$R$ (mask ratio) & 0.8 & 0.75& 0.75 \\
$A$ (mask adjust) & 0.07 & 0.1& 0.1 \\
$K$ (layers to average) & 10 & 18& 32 \\
Target normalization & IN $\rightarrow$ AVG $\rightarrow$ LN & IN $\rightarrow$ AVG $\rightarrow$ LN& IN $\rightarrow$ AVG $\rightarrow$ LN \\
Updates & 500,000 & 750,000& 500,000 \\
Decoder dim. & 768 & 1024& 1024 \\
Decoder conv. groups & 16 & 16& 16 \\
Decoder kernel & 3 & 5& 5 \\
Decoder layers ($D$) & 6 & 3& 3 \\
\bottomrule
\end{tabular}
\end{table}
}

\newcommand{\insertHyperparameterTableSpeech}{
\begin{table}[h!]
\centering
\caption{Speech pre-training hyper-parameters. IN is instance normalization; AVG is mean pooling.
\label{tab:speech_hyperparams}}
\begin{tabular}[t]{lrr}
\toprule
& Base (Librispeech) & Large (Libri-light) \\
\midrule
GPUs & 16 & 64 \\
Learning rate & $\Enot{7.5e-4}$ & $\Enot{4e-4}$ \\
Adam $\beta_1$ / $\beta_2$ & 0.9 / 0.98 & 0.9 / 0.98 \\
Weight decay & 0.01 & 0.01 \\
Clip norm & - & 1 \\
Learning rate schedule & cosine & cosine \\
Warmup updates & 8,000 & 10,000 \\
Batch size (seconds per GPU / total) & 62.5 / 1,000 & 20 / 960 \\
Multi-masks ($M$) & 8 & 12 \\
$\tau_0$ (EMA start) & 0.999 & 0.9997 \\ 
$\tau_e$ (EMA end) & 0.99999 & 1.0 \\ 
$\tau_n$ (EMA anneal steps) & 75,000 & 300,000 \\ 
$B$ (block width) & 5 & 5 \\
$R$ (mask ratio) & 0.5 & 0.55 \\
$A$ (mask adjust) & 0.05 & 0.1 \\
$K$ (layers to average) & 8 & 16 \\
Target normalization & IN $\rightarrow$ AVG & IN $\rightarrow$ AVG \\
Updates & 400,000 & 600,000 \\
Decoder dim. & 384 & 768 \\
Decoder conv. groups & 16 & 16 \\
Decoder kernel & 7 & 7 \\
Decoder layers ($D$) & 4 & 4 \\
\bottomrule
\end{tabular}
\end{table}
}

\newcommand{\insertHyperparameterTableNLP}{
\begin{table}[h!]
\centering
\caption{Natural language processing pre-training hyper-parameters. IN is instance normalization; AVG is mean pooling.
\label{tab:nlp_hyperparams}}
\begin{tabular}[t]{lr}
\toprule
& Base  \\
\midrule
GPUs & 16  \\
Learning rate & $\Enot{2e-4}$ \\
Adam $\beta_1$ / $\beta_2$ & 0.9 / 0.98 \\
Weight decay & 0.01 \\
Clip norm & 1.0 \\
Learning rate schedule & cosine  \\
Warmup updates & 4,000  \\
Batch size & 32 \\
Multi-masks ($M$) & 8 \\
$\tau_0$ (EMA start) & 0.9999  \\ 
$\tau_e$ (EMA end) & 1 \\ 
$\tau_n$ (EMA anneal steps) & 100,000  \\ 
$B$ (block width) & 1 \\
$R$ (mask ratio) & 0.42  \\
$A$ (mask adjust) & 0 \\
$K$ (layers to average) & 12  \\
Target normalization & IN $\rightarrow$ AVG \\
Updates & 1,000,000 \\
Decoder dim. & 768 \\
Decoder conv. groups & 1 \\
Decoder kernel & 9 \\
Decoder layers ($D$) & 5  \\
\bottomrule
\end{tabular}
\end{table}
}

\icmltitlerunning{Efficient Self-supervised Learning with Contextualized Target Representations for Vision, Speech and Language}

\begin{document}

\twocolumn[
\icmltitle{Efficient Self-supervised Learning with Contextualized Target Representations for Vision, Speech and Language}



\icmlsetsymbol{equal}{*}

\begin{icmlauthorlist}
\icmlauthor{Alexei Baevski}{char}
\icmlauthor{Arun Babu}{meta}
\icmlauthor{Wei-Ning Hsu}{meta}
\icmlauthor{Michael Auli}{meta}
\end{icmlauthorlist}

\icmlaffiliation{char}{Character AI, work done while at Meta AI}
\icmlaffiliation{meta}{Meta AI}

\icmlcorrespondingauthor{Alexei Baevski}{alexei.b@gmail.com }
\icmlcorrespondingauthor{Michael Auli}{michaelauli@meta.com}

\icmlkeywords{Self-supervised Learning, Deep Learning}

\vskip 0.3in
]



\printAffiliationsAndNotice{}  % leave blank if no need to mention equal contribution

\begin{abstract}
Current self-supervised learning algorithms are often modality-specific and require large amounts of computational resources. 
To address these issues, we increase the training efficiency of data2vec, a learning objective that generalizes across several modalities.
We do not encode masked tokens, use a fast convolutional decoder and amortize the effort to build teacher representations.
\name{} benefits from the rich contextualized target representations introduced in data2vec which enable a fast self-supervised learner.
Experiments on ImageNet-1K image classification show that \name{} matches the accuracy of Masked Autoencoders in 16.4x lower pre-training time, on Librispeech speech recognition it performs as well as wav2vec 2.0 in 10.6x less time, and on GLUE natural language understanding it matches a retrained RoBERTa model in half the time.
Trading some speed for accuracy results in ImageNet-1K top-1 accuracy of 86.8\% with a ViT-L model trained for 150 epochs.
Models and code are available at
{\small\url{www.github.com/pytorch/fairseq/tree/master/examples/data2vec}}.
\end{abstract}

\section{Introduction}

Self-supervised learning has been an active research topic which resulted in much progress across several areas such as computer vision~\citep{grill2020byol,bao2021beit,he2021mae}, natural language processing~(NLP;~\citealt{radford2018unsup,devlin2018bert,raffel2019t5,brown2020gpt3}), and speech processing~\citep{oord2018cpc,schneider2019wav2vec,baevski2020wav,hsu2020hubert,baevski2021unsupervised,babu2022xlsr}.
However, algorithms are often designed with a single modality in mind which makes it unclear whether the same learning mechanisms generalize across modalities.
To this end, recent work has introduced unified model architectures~\citep{jaegle2021perceiver,jaegle2022perceiverio} and training objectives which function identically in different modalities~\citep{baevski2022d2v}.

Self-supervised models have benefited from increased scale in model capacity and training datasets~\citep{brown2020gpt3} as well as large amounts of computational training effort~\citep{hoffmann2022chinchilla} which resulted in interesting emerging properties~\citep{wei2022emergent}.
And while the resulting models are excellent few-shot learners~\citep{brown2020gpt3}, the preceding self-supervised learning stage is far from efficient: for some modalities,
models with hundreds of billions of parameters are trained which often pushes the boundaries of what is computationally feasible. 

\begin{figure*}[h!t]
\begin{center}
\includegraphics[width=1\linewidth]{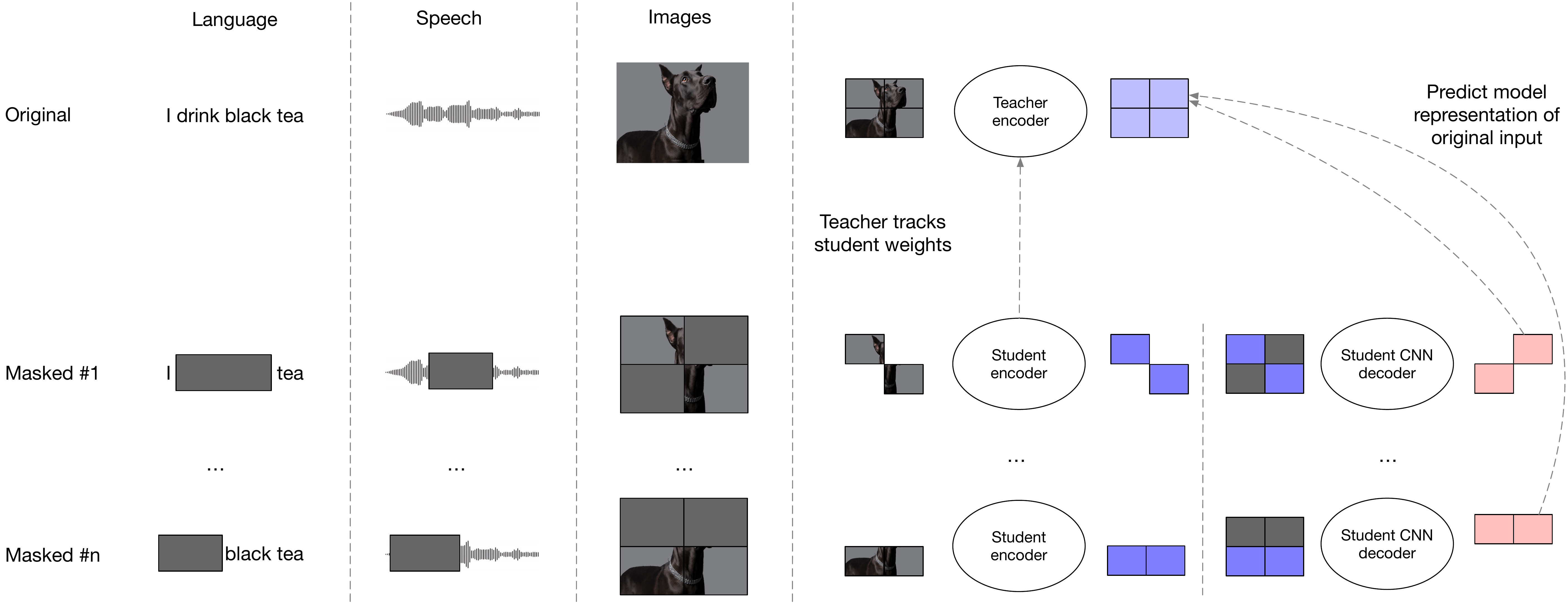}
\caption{\name{} uses the same learning objective for different modalities (but trains different models for each). 
We first create a \emph{contextualized target} representation based on the unmasked training sample using the teacher model whose weights are an exponentially moving average of the student model. 
Target representations are contextualized due to self-attention in Transformer models.
The same target representation is predicted by the student model for different masked versions of the training example, thereby amortizing the computational cost of creating target representations.
Masked portions of the training sample are not encoded~\citep{he2021mae}.
\label{fig:model}
}
\end{center}
\end{figure*}

In this paper, we present \name{} which improves the compute efficiency of self-supervised learning with  contextualized target prediction~\citep{baevski2022d2v} by using an efficient data encoding~\citep{he2021mae}, a fast convolutional decoder and by reusing target representations for multiple masked versions of each sample~\citep{assran2022masked,girdhar2022omnimae}. 
The algorithm uses the same learning objective for each modality but trains separate models for each one using the Transformer architecture with different feature encoders depending on the input modality.
We follow~\citet{baevski2022d2v} by creating latent contextualized representations with a teacher model based on unmasked training examples which are regressed by a student model whose input is a masked version of the sample (\autoref{fig:model}) 

Target contextualization enables capturing information about the entire sample, e.g., for text, these targets can represent the different meanings of a word depending on the context.
This is more difficult for conventional non-contextualized targets which use a single set of features to represent the different meanings of a word.
At first glance, the creation of contextualized targets with a separate teacher appear to be an additional step that slows model training but our efficiency improvements suggest that contextualized targets result in a richer learning task and faster learning.

Experiments demonstrate efficiency improvements of between 2-16x at similar accuracy on image classification, speech recognition and natural language understanding.

\section{Related Work}

\paragraph{Self-supervised learning for NLP, Speech and Vision.}
There has been much work on self-supervised learning for individual modalities such as NLP where models segment text into sub-word units and define the learning task based on these units by predicting either the next token for causal models~\citep{radford2018unsup,brown2020gpt3,chowdhery2022palm} or by predicting masked tokens for bi-directional models~\citep{devlin2018bert,baevski2019bitransformer}.

For speech processing, models either reconstruct the audio signal~\citep{eloff2019unsupervised,liu2021tera} or solve a learning task based on discretizing  short and overlapping windows of the speech signal, either in a left-to-right fashion~\citep{oord2018cpc,schneider2019wav2vec,baevski2019vqwav2vec} or using masked prediction~\citep{baevski2020wav,hsu2020hubert,chen2021wavlm}.

In computer vision, there has been a shift towards Vision Transformer architectures (ViT;~\citealt{dosovitskiy2020vit}) and masked prediction methods that can be very efficient by not encoding masked patches~\citep{he2021mae}. 
There are also approaches that learn based on discrete visual tokens~\citep{bao2021beit,peng2022beit2}.
Other approaches are based on self-distillation~\citep{grill2020byol,caron2021dino} and online clustering~\citep{caron2020swav}.

Related work to our multi-mask training regime (\textsection\ref{sec:multimask}) includes \citet{caron2020contrastive} which creates multiple crops from the same image which contrasts to our approach of creating multiple masked versions of the same training example.
\citet{jing2022maskedsiamese} also experiment with applying different masks in the context of convolutional neural networks to a training example but find that data augmentations such as cropping and flipping outperform this masking strategy.
Finally, \citet{wu2022extreme} also consider multiple masks but they predict representations of the entire training sample and do not average multiple layers.

\paragraph{Generalized Architectures and Learning Objectives.}
Another trend is the unification of neural network architectures that can process data from different modalities using the same network~\citep{jaegle2021perceiver,jaegle2022perceiverio}. 
This is complemented by work on unifying the self-supervised learning objective for vision, speech, and text in data2vec~\citep{baevski2022d2v}. 
A distinguishing characteristic of data2vec is that it is trained by predicting contextualized target representations which contain features from the entire input example compared to the limited information of a particular time-step or patch.

\paragraph{Joint Multi-modal Learning.}
While data2vec and the current work are trained for each modality individually, there has been considerable work on training joint modality models which can represent multiple modalities within the same model. 
This includes models trained on images and text~\citep{radford2021clip,singh2021flava,wang2021vlmo,alayrac2022flamingo}, speech and text~\citep{shi2022avhubert}, or video/audio/text~\citep{akbari2021vatt}.

\paragraph{Efficient Self-supervised Learning.}
After the success of BERT in NLP, follow on work include a more lightweight training objective to increase efficiency~\citep{clark2020electra} and work on reducing the model capacity through weight sharing~\citep{lan2019albert} which resulted in faster training speed.
In computer vision, \citet{he2021mae} introduced the idea of not processing masked patches in the encoder network which increased training speed and \citet{assran2022masked} used this idea in a joint embedding architecture to achieve label efficient self-supervised learning. 
There is also work on sparse attention to increase efficiency~\citep{li2021esvit}.
For speech, more efficient feature encoder models and time-step squeezing has helped to improve efficiency~\citep{wu2022interspeech,vyas2022interspeech}.

\section{Method}

Our approach builds on data2vec~\citep{baevski2022d2v} and we first describe the major shared techniques including predicting contextualized target representations (\textsection\ref{sec:d2v}).
Similar to Masked Autoencoders (MAE;~\citealt{he2021mae}), we encode only non-masked portions of a sample and use a decoder model to predict target representations for the masked portions but 
instead of using a Transformer-based decoder, we use a smaller convolutional decoder which we find to be easier and faster to train (\textsection\ref{sec:architecture}).
To amortize the computational overhead of creating contextualized target representations, we reuse each target for multiple masked versions of a training sample (\textsection\ref{sec:multimask}) and instead of random masking or block masking, our inverse block masking strategy ensures that contiguous regions of the sample are preserved to provide more context for student predictions (\textsection\ref{sec:masking}).

\subsection{Contextualized Target Prediction}
\label{sec:d2v}

Instead of reconstructing local windows of the the raw input data~\citep{he2021mae}, or predicting discrete representations thereof~\citep{bao2021beit}, we predict representations of the teacher network incorporating information from the entire input sample. 
This leads to a richer training task where targets are specific to a particular training sample.
Contextualized targets are built via the self-attention mechanism of a Transformer-based teacher model which encodes the unmasked training sample~\citep{paulus17intra,vaswani2017transformer} and the training targets are a weighted sum of all features in the sample.

\paragraph{Target Representations and Learning Objective.}
Training targets are based on averaging the top $K$ FFN blocks of the teacher. 
Before averaging, activations are normalized using instance normalization~\citep{ulyanov2016in}.\footnote{Layer normalization~\citep{ba2016layer} of the averaged targets can be useful for some modalities such as speech and vision.}
The training task is for the student network to regress these targets based on the masked version of the sample.

\paragraph{Teacher Weights.}
The teacher weights $\Delta$ are an exponentially moving average of the student encoder weights $\theta$ \citep{grill2020byol}: $\Delta \leftarrow \tau \Delta + (1 - \tau)~\theta$ where $\tau$ follows a linearly increasing schedule from a starting value $\tau_0$ to a final value $\tau_e$ over $\tau_n$ updates, after which the value is kept constant~\citep{baevski2022d2v}.

\paragraph{Learning Objective.}
We use an L2 loss based on the target representation from the teacher network $\mathbf{y}$ and the student network prediction $\mathbf{f(x)}$.
This is a simplification compared to the Smooth L1 loss used in~\citet{baevski2022d2v} and we found it to work well across modalities.

\subsection{Model Architecture}
\label{sec:architecture}

% \paragraph{Transformer Backbone and Modality-specific Feature Encoders.}
Similar to data2vec~\citep{baevski2022d2v}, our model uses modality-specific feature encoders and a Transformer architecture where the latter makes up the the bulk of the model weights~\citep{vaswani2017transformer}.
For computer vision, we use a patch mapping of 16x16 pixels as feature encoder~\citep{dosovitskiy2020vit}, for speech a multi-layer convolutional network following~\citet{oord2018cpc,baevski2020wav,baevski2022d2v} and for text we use embeddings learned based on byte-pair encoding~\citep{sennrich2016bpe}.

\paragraph{Asymmetric Encoder/Decoder Architecture.}
In a first step, we use the teacher network to encode all parts of the unmasked training sample in order to create training targets (\textsection\ref{sec:d2v}).
Next, we mask part of the sample (\textsection\ref{sec:masking}) and embed it with the student encoder. 
To improve efficiency, we encode only unmasked patches or time-steps of a training example which leads to a large speed-up compared to encoding all parts of the sample~\citep{he2021mae}, depending on the amount of masking.
The output of the student encoder is then merged with fixed representations for the masked portions and fed to a decoder network.
To represent the masked tokens, we found it sufficient to use random Gaussian noise compared to a learned representation~\citep{he2021mae}.\footnote{We also experimented with adding positional embeddings but found that they do not improve results.}
The decoder network then reconstructs the contextualized target representation of the teacher network for time-steps which are masked in the student input.

\paragraph{Convolutional Decoder Network.}
We use a lightweight decoder consisting of $D$ convolutions, each followed by layer normalization~\citep{ba2016layer}, a GELU activation function~\citep{hendrycks2016gaussian}, and a residual connection~\citep{he2015deep}.
For sequential data such as speech and text we use 1-D convolutions and for images we use 2-D convolutions, each parameterized by groups to increase efficiency~\citep{krizhevsky2012alexnet}.
We tune the number of layers and kernel size for each modality.

\subsection{Multi-mask Training}
\label{sec:multimask}

A disadvantage of the data2vec teacher-student setup is the need to process each sample twice: once to obtain targets with the teacher model, and once to obtain predictions of the student.
Moreover, computing activations for the teacher model is also less efficient compared to the student model since the teacher needs to process the full unmasked input.\footnote{\citet{baevski2022d2v} found it important to build targets based on the unmasked sample rather than another masked version.} 

In order to amortize the cost of the teacher model computation, we reuse the teacher representation for multiple masked versions of the training sample.
Concretely, we consider $M$ different masked versions of the training sample and compute the loss with respect to the same target representation. 
This is possible, because target representations are based on the full unmasked version of the sample.
As $M$ grows, the computational overhead of computing target representations becomes negligible.
In practice, this enables training with a relatively small batch size compared to other self-supervised work (\textsection\ref{sec:results}).

Considering multiple masked versions of a training sample has been previously explored in the context of self-supervised learning for computer vision with ResNet models~\citep{jing2022maskedsiamese}, although the authors found that it performed much less well than different image augmentations. \citet{caron2020contrastive} considers multiple crops based on the same image but trains the model by comparing discrete codes rather than predicting the representation of the original image.
And~\citet{girdhar2022omnimae} trains MAE models on videos with multiple masked versions of a sample to amortize the overhead of data loading and preparation.

Another efficiency improvement of \name{} compared to data2vec is to share the feature encoder output across the different masked versions of the training example to avoid redundant computation.
This leads to significant speed-ups for dense modalities such as speech where the feature encoder accounts for a large portion of the computation but less so for other modalities such as text.

\subsection{Inverse Block Masking}
\label{sec:masking}

The MAE-style sample encoding improves efficiency but also removes the ability to store information in the activations of masked time-steps which makes the training task more challenging.
Random masking is successful for Masked Autoencoders~\citep{he2021mae} but it may interfere with the ability to build semantic representations since there is no structure in the masks that are created.
Block masking~\citep{bao2021beit} is more structured by masking entire blocks of time-steps or patches but there is no guarantee that large contiguous portions of the training sample are unmasked.
Our goal is to enable the student model to build semantically rich representations over local regions of the sample.

\insertEfficiencyPlot

We therefore introduce inverse block masking: instead of choosing which patches to mask, it chooses which patches to preserve in a block-wise fashion, where the size of a block is in terms of the number of patches or time-steps $B$.
We first sample the starting point of each block to keep, and then expand it symmetrically until the block is of width $B$, for speech and text, or $\sqrt{B}$ for images.\footnote{For speech and text the blocks are 1-D and block masking/inverse block masking perform similarly due to symmetry.}
We sample the following number of starting points without replacement and expand them to width $B$ or quadratic blocks of width $\sqrt{B}$, depending on the modality: 
$$
L \times \frac{(1 - R) + A}{B}
$$
where $L$ is the total number of time-steps/patches in a training sample, $R$ is the mask ratio, a hyper parameter controlling the percentage of the sample that is masked and $A$ is a hyper-parameter to adjust mask ratio (see below).

We allow blocks to overlap, which results in over-masking and some variance in the number of actually masked time-steps for each sample.
Since we only encode unmasked time-steps, we use a simple strategy to assimilate the number of unmasked time-steps for all samples in a batch:
for each sample, we randomly choose individual time-steps to mask or unmask until we reached the desired number of unmasked time-steps $L \times (1 - R)$.\footnote{We found $0.05 < A < 0.15$ to work well.}

\section{Experiments}
\label{sec:results}

\subsection{Efficiency}

As a first experiment, we compare the efficiency of \name{} pre-training to existing algorithms for vision, speech and NLP. 
We measure accuracy for image classification~(\textsection\ref{sec:cv}), word error rate for speech recognition (\textsection\ref{sec:speech}), natural language understanding performance on GLUE~(\textsection\ref{sec:nlp}) and pre-training speed in terms of wall clock hours.

\paragraph{Setup.}
For computer vision, we compare to MAE~\citep{he2021mae} and data2vec~\citep{baevski2020wav} using their public implementations and recommended configurations.\footnote{data2vec: {\scriptsize\url{https://github.com/facebookresearch/fairseq/tree/main/examples/data2vec}} MAE: {\scriptsize\url{https://github.com/facebookresearch/mae/blob/main/PRETRAIN.md}}
}
Both \name{} and data2vec are implemented in fairseq~\citep{ott2019fairseq} and we evaluate \name{} configurations with different speed and accuracy trade-offs.
All vision models are pre-trained on 32 A100 40GB GPUs, \name{} models are pre-trained between 25k-500k updates, or 10-200 epochs, all with a total batch size of 512 images, $R=0.8$, $M=8$, except for the longest training run which uses $M=16$.
MAE is pre-trained for 500k updates using a batch size of 4,096 images or 1,600 epochs; data2vec is pre-trained for 500k updates with batch size 2,048 or 800 epochs.

Speech models are pre-trained on 32 A100 40GB GPUs, and \name{} performs between 50k-400k updates, or 13-103 epochs, using a total batch size of 17 minutes of speech audio and we set $M=8$, $R=0.5$. 
We compare to wav2vec 2.0 and data2vec which are pre-trained for 400k updates with batch size 93min and 63min, respectively and following the recommended configurations. 
All models are implemented in fairseq.

NLP models are pre-trained on 16 A100 40GB GPUs, \name{} uses between 400k-1m updates with a total batch size of 32 (each sample is a sequence of 512 tokens) and we set $M=8$, $R=0.42$.
Models are compared to a retrained version of RoBERTa and data2vec are both pre-trained for 1m updates with a total batch size of 256 (32 epochs) following the original BERT setup.
Models are implemented in fairseq.

\paragraph{Results.}
\autoref{fig:efficiency} shows that \name{} provides a far better speed and accuracy trade-off in all three modalities: 
an ImageNet pre-trained \name{} model achieves a top-1 accuracy of 83.7\% after pre-training for just over 3 hours vs. 83.6\% after 50.7 hours for MAE - a 16.4x speed-up at slightly improved accuracy compared to the popular MAE algorithm~\citep{he2021mae}.
\ma{This is equivalent to 99 hours on a single A100 GPU, and probably less due to multi-GPU training overhead.}
A speech \name{} model achieves comparable word error rate to wav2vec 2.0 on speech recognition in 10.6x times lower wall clock time.
For NLP, \name{} trains to a similar accuracy as a retrained RoBERTa model in two times the speed. 

The same models also perform far fewer epochs: for computer vision the \name{} model with most similar accuracy to MAE performs 20 epochs vs. 1,600 epochs.
For speech, \name{} trains for 13 epochs vs. 522 epochs and for NLP, \name{} performs four epochs compared to 32 for RoBERTa.
\name{} also provides a better efficiency compared to data2vec~\citep{baevski2022d2v}: for vision, \name{} can nearly match the accuracy of data2vec in 2.9x less time, for speech there is 3.8x speed-up, and for NLP there is 2.5x speed-up.

Note that data2vec is already faster than MAE: 
the most comparable data2vec model trained in 2.8x the speed of MAE (17.8 hours vs. 50.7 hours) - at higher accuracy (84.0\% vs. 83.6\%). 
Hence, the speed-up of \name{} compared to MAE is much higher than for NLP, where the original data2vec was not more efficient than RoBERTa.

\name{} can train well with a relatively small batch size of just 512 images, compared to 4,096 images in the case of MAE, or 2,048 images for data2vec and most other self-supervised algorithms for computer vision (\textsection\ref{sec:ablations} analyzes multi-masking in more detail).
Training with a much lower number of epochs and batch size is possible because multi-masking extracts more learning signal from each training sample.
Moreover, contextualized targets lead to a richer training task.

\subsection{Computer Vision}
\label{sec:cv}

Next, we compare \name{} more broadly for each modality to existing work.
For computer vision, we use a standard Vision Transformer architecture~\citep{dosovitskiy2020vit} but with post-layer normalization, similar to the original Transformer architecture~\citep{vaswani2017transformer}.
This results in an identical Transformer architecture for all  modalities.
We also apply random cropping and horizontal flipping to input images whose result and we feed the same augmented version both to the student and the teacher; we use the same hyper-parameters as MAE~\citep{he2021mae}.
For vision only, we found it useful to add a global CLS loss~\citep{peng2022beit2}.
For detailed hyper-parameters see~\autoref{appendix:hyper} \autoref{tab:imagenet_hyperparams}; for fine-tuning we use the same settings as~\citet{he2021mae}.
We pre-train on the unlabeled version of ImageNet-1K. 

\autoref{tab:in} shows that \name{} improves over prior single models using no external data both for ViT-B and ViT-L while training for far fewer epochs: 
compared to MAE, \name{} increases accuracy by 0.9\% while pre-training for less time (ViT-B: 32 hours vs. 50.7 hours, ViT-L: 63.3 hours vs. 93.3 hours).
Compared to data2vec~\citep{baevski2022d2v} achieves slightly higher accuracy at far fewer epochs.
\name{} also improves over several approaches using multiple models and/or external data such as TEC~\citep{gao2022sustainssl}, PeCo~\citep{dong2022peco}, and BEiT~\citep{bao2021beit}. 
BEiT-2~\citep{peng2022beit2} performs better because it effectively distills representations from CLIP~\citep{radford2021clip} which was trained on a much larger dataset than ImageNet-1K.

\insertINtable

\autoref{tab:inhuge} shows the speed/accuracy trade-off for ViT-H models: \name{} outperforms MAE by 0.5\% while training for 40\% less time and performing 1/16 of the number of training epochs.

\subsection{Speech Processing}
\label{sec:speech}

\insertLLtable
To evaluate \name{} on speech, we pretrain it on either Librispeech~\citep{panayotov2015librispeech} or the much larger Libri-light dataset~\citep{librilight} and fine-tune the resulting model for speech recognition on the labeled data splits of Libri-light which tests the model quality for different resource settings.
See~\autoref{tab:speech_hyperparams} in Appendix \ref{appendix:hyper} for detailed hyper-parameters.
We follow the fine-tuning regime of wav2vec 2.0~\citep{baevski2020wav} whose hyper-parameters depend on the labeled data setup.

\paragraph{Alibi feature encoder.}
The feature encoder of~\citet{baevski2020wav} uses relative positional embeddings modeled as a temporal convolution which assumes the that all time-steps are being encoded. 
We adapt this to our setup by removing parts of the kernel corresponding to masked time-steps.
We also found it helpful to bias the query-key attention scores with a penalty proportional to their distance~\citep{press2021alibi}.
Biases are initialized following~\citet{press2021alibi}, but we keep them frozen during training and learn a scalar for each head which is initialized to $1.0$. 
This adds very few new parameters ($16$ for a Large model), but leads to a significant improvement in accuracy which we ablate in~\textsection\ref{sec:ablations}.

The results (\autoref{tab:librilight}) show that \name{} improves in most settings over prior work in less training time. 
Compared to wav2vec 2.0, \name{} enables a relative word error rate reduction of up to 26\% for Base models and up to 18\% for Large models.
For Base models, we use the most accurate model of~\autoref{fig:efficiency_vision} which obtains higher accuracy than other models at faster training time (43.3 hours on 16 A100 40GB GPUs which as wav2vec 2.0 requires 57.3 hours on the same hardware).
For Large models, we train \name{} on 64 A100 40GB GPUs for 76.7 hours while as other models train for either 108 hours (data2vec) or 150 hours (wav2vec 2.0) on the same hardware.

\subsection{Natural Language Processing}
\label{sec:nlp}

\insertGLUEtable

For NLP, we adopt the same training setup as BERT~\citep{devlin2018bert} by pre-training on the Books Corpus~\citep{zhu2015books} and English Wikipedia using a 50k byte-pair encoding~\citep{sennrich2016bpe,devlin2018bert,liu2019roberta}.
As baseline we retrain RoBERTa using the original BERT setup~(Baseline;~\citealt{liu2019roberta}) with the default BERT masking strategy (mask 15\% of tokens) but without the next-sentence prediction task and we also compare to data2vec. 
Both RoBERTa and data2vec are pre-trained for 1m updates and with batch size 256.
The hyper-parameters for \name{} are in~\autoref{appendix:hyper} \autoref{tab:nlp_hyperparams}.

Models are evaluated on the General Language Understanding Evaluation (GLUE) benchmark~\citep{wang2018glue} comprising tasks for natural language inference (MNLI, QNLI, RTE), sentence similarity (MRPC, QQP and STS-B), grammaticality (CoLA), and sentiment analysis (SST-2).
% \footnote{MNLI (Multi Genre Natural Language Inference;~\citealt{williams2017mnli}), Stanford Question Answering Dataset (QNLI;~\citealt{rajpurkar2016squad}), Recognizing Textual Entailment~(RTE;\citealt{dagan2006rte1,haim2006rte2,giampiccolo2007rte3,bentivogli2009rte}), and we exclude Winograd NLI task from our results similar to~\citet{devlin2018bert},
% Microsoft Research Paragraph Corpus (MRPC; \citealt{dolan2005mrpc}), Quora Question Pairs benchmark (QQP), and the Semantic Textual Similarity Benchmark (STS-B; \citealt{cer2017stsb}),
% Corpus of Linguistic Acceptability (CoLA;~\citealt{warstadt2018cola}),
% Stanford Sentiment Treebank (SST-2;~\citealt{socher2013sst2})}
Pre-trained models are fine-tuned on the labeled data provided by each task and we report the average accuracy on the development sets by performing nine different fine-tuning runs and reporting the average performance without the two best and the two worst performing runs to reduce the sensitivity to outliers.

The results (\autoref{tab:glue}) shows that \name{} achieves comparable average GLUE performance to our retrained RoBERTa baseline in 1.8x the speed and 7.8 fewer epochs.
Compared to data2vec, there is a 2.5x speed-up.
Note that \name{} uses a much higher masking rate of 42\% compared to 15\% for BERT/RoBERTa which we believe is possible due to the use of rich contextualized targets.

\subsection{Ablations}
\label{sec:ablations}

\insertBszCloneAblationPlot

\insertAblationTable

\paragraph{Multi-mask training.}
Next, we analyze the effect of multi-masking for different batch sizes. 
We use a reduced computer vision setup where we pre-train with 100k updates for a given batch size (bsz). 
\autoref{fig:multimask} shows that considering multiple masks per training sample can drastically improve  accuracy, e.g., for bsz=64 considering $M=16$ instead of $M=2$ raises accuracy by 4.6\% keeping everything else equal.
This effect decreases with larger batch sizes but shows the possibility of pre-training high-quality models with dramatically lower batch size than is common today.

\paragraph{Training Losses.}
In the next experiment, we study our loss in more detail. 
\autoref{tab:losses} shows that the CLS loss component (\textsection\ref{sec:cv}) leads to a small improvement in accuracy for computer vision.
The prediction of global representations as done by the CLS loss is complementary to predicting local patch information.
We also compare contextualized target prediction to regressing the raw pixels of a local 16x16 patch of the training sample~\citep{he2021mae}.
Adding the MAE pixel regression loss (pixel regr) does not improve over contextualized target prediction alone and training only with the pixel regression loss (pixel regr only) results in a substantial drop in accuracy.

\paragraph{Masking Strategy.}
Next, we ablate our masking strategy by comparing it to block masking and random masking. 
\autoref{tab:masking} shows that block masking (block) performs less well than inverse block masking (our standard setting is $B=3$); $B=1$ corresponds to random masking and is also less effective.

\paragraph{Speech Alibi Embeddings.}
Finally, we investigate the effectiveness of the relative position embeddings for speech.
\autoref{tab:alibi} shows that the convolutional embeddings alone (baseline - alibi) perform less well than the alibi embeddings and that our design choices of learning scalars for the random embeddings are effective.

\section{Conclusion and Future Work}
We presented an efficient and general pre-training technique which relies on the same learning objective in different modalities. 
\name{} shows that the training speed of self-supervised learning can be substantially improved with no loss in downstream task accuracy.
At the heart of our approach lies the use of contextualized target representations with result in a more efficient self-supervised learner.
Experiments show that \name{} can reach the same accuracy as many popular existing algorithms in 2-16x the training speed.
Future work includes the application of \name{} to other modalities than vision, speech and text.

% % Acknowledgements should only appear in the accepted version.
% \section*{Acknowledgements}

% We thank Brenden Lake and Emmanuel Dupoux for helpful discussions. 

% \textbf{Do not} include acknowledgements in the initial version of
% the paper submitted for blind review.

% If a paper is accepted, the final camera-ready version can (and
% probably should) include acknowledgements. In this case, please
% place such acknowledgements in an unnumbered section at the
% end of the paper. Typically, this will include thanks to reviewers
% who gave useful comments, to colleagues who contributed to the ideas,
% and to funding agencies and corporate sponsors that provided financial


\bibliography{refs}
\bibliographystyle{icml2022}

%%%%%%%%%%%%%%%%%%%%%%%%%%%%%%%%%%%%%%%%%%%%%%%%%%%%%%%%%%%%%%%%%%%%%%%%%%%%%%%
%%%%%%%%%%%%%%%%%%%%%%%%%%%%%%%%%%%%%%%%%%%%%%%%%%%%%%%%%%%%%%%%%%%%%%%%%%%%%%%
% APPENDIX
%%%%%%%%%%%%%%%%%%%%%%%%%%%%%%%%%%%%%%%%%%%%%%%%%%%%%%%%%%%%%%%%%%%%%%%%%%%%%%%
%%%%%%%%%%%%%%%%%%%%%%%%%%%%%%%%%%%%%%%%%%%%%%%%%%%%%%%%%%%%%%%%%%%%%%%%%%%%%%%
\newpage
\appendix
\onecolumn

\section{Pre-training Hyper-parameters}
\label{appendix:hyper}

\insertHyperparameterTableVision
\insertHyperparameterTableSpeech
\insertHyperparameterTableNLP

\clearpage

\section{Effect of Pre-training Dataset Size}

Figure \ref{fig:datasetsize} shows the effect of randomly subsampling the pre-training data while keeping all hyper-parameters and the fine-tuning data constant. Increasing the amount of pre-training data helps larger models more, implying that the base size models underfit to ImageNet-1K with the data2vec style pre-training task.

\insertDatasetSzPlot

\end{document}